\documentclass[runningheads]{llncs}
\usepackage{graphicx}
\usepackage{amsmath,amssymb} 
\usepackage{color}
\usepackage{url}

\usepackage{multirow}
\usepackage{array}

\graphicspath{{./}{./figure/}}
\DeclareGraphicsExtensions{.pdf,.jpeg,.png,.jpg}

\makeatletter
\DeclareRobustCommand\onedot{\futurelet\@let@token\@onedot}
\def\@onedot{\ifx\@let@token.\else.\null\fi\xspace}

\def\Mat#1{{\boldsymbol{#1}}}

\begin{document}
\title{Deep Feature Pyramid Reconfiguration for Object Detection}

\titlerunning{Deep Feature Pyramid Reconfiguration for Object Detection}
%
\author{Tao Kong\inst{1} \and
Fuchun Sun \inst{1}  \and
Wenbing Huang \inst{2} \and
Huaping Liu \inst{1}
}
%
\authorrunning{Tao Kong et al.}
%

\institute{ Department of Computer Science and Technology, Tsinghua University,\\
Beijing National Research Center for Information Science and Technology (BNRist).\\
\email{\{kt14@mails, fcsun@mail, hpliu@mail\}.tsinghua.edu.cn}
\and
Tencent AI Lab. \\
\email{hwenbing@126.com}\\}
\maketitle              
\begin{abstract}
State-of-the-art object detectors usually learn multi-scale representations to get better results by employing feature pyramids.
However, the current designs for feature pyramids are still inefficient to integrate the semantic information over different scales.
In this paper, we begin by investigating current feature pyramids solutions, and then reformulate the feature pyramid construction as the feature reconfiguration process.
Finally, we propose a novel reconfiguration architecture to combine low-level representations with high-level semantic features in a highly-nonlinear yet efficient way.
In particular, our architecture which consists of global attention and local reconfigurations, is able to gather task-oriented features across different spatial locations and scales, globally and locally.
Both the global attention and local reconfiguration are lightweight, in-place, and end-to-end trainable.
Using this method in the basic SSD system, our models achieve consistent and significant boosts compared with the original model and its other variations, without losing real-time processing speed.
\end{abstract}

\section{Introduction}

Detecting objects at vastly different scales from images is a fundamental challenge in computer vision \cite{pyramid_intro}. One traditional way to solve this issue is to build feature pyramids upon image pyramids directly. Despite the inefficiency, this kind of approaches have been applied for object detection and many other tasks along with hand-engineered features \cite{fast-fp,dpm_fp}.

We focus on detecting objects with deep ConvNets in this paper. Aside from being capable of representing higher-level semantics, ConvNets are also robust to variance in scale, thus making it possible to detect multi-scale objects from features computed on a single scale input \cite{fasterrcnn,fastrcnn}. However, recent works suggest that taking pyramidal representations into account can further boost the detection performance \cite{fpn,sppnet,mrcnn}. This is due to its principle advantage of producing multi-scale feature representations in which all levels are semantically strong, including the high-resolution features.

There are several typical works exploring the feature pyramid representations for object detection. The Single Shot Detector (SSD) \cite{ssd} is one of the first attempts on using such technique in ConvNets. Given one input image, SSD combines the predictions from multiple feature layers with different resolutions to naturally handle objects of various sizes.
However, SSD fails to capture deep semantics for shallow-layer feature maps, since the bottom-up pathway in SSD can learn strong features only for deep layers but not for the shallow ones.
This causes the key bottleneck of SSD for detecting small instances.

To overcome the disadvantage of SSD and make the networks more robust to object scales, recent works (e.g., FPN \cite{fpn}, DSSD \cite{dssd}, RON \cite{ron} and TDM \cite{tdm}) propose to combine low-resolution and semantically-strong features with high-resolution and semantically-weak features via lateral connections in a top-down pathway.  In contrast to the bottom-up fashion in SSD, the lateral connections pass the semantic information down to the shallow layers one by one, thus enhancing the detection ability of shallow-layer features. Such technology is successfully used in object detection \cite{dssd,focal}, segmentation \cite{maskrcnn}, pose estimation \cite{fpn_pose,cpn}, etc.

Ideally, the pyramid features in ConvNets should: (1) reuse multi-scale features from different layers of a single network, and (2) improve features with strong semantics at all scales.
The FPN works \cite{fpn} satisfy these conditions by lateral connections. Nevertheless, the FPN, as demonstrated by our analysis in \textsection~\ref{Sec:GLR}, is actually equivalent to a linear combination of the feature hierarchy. Yet, the linear combination of features is too simple to capture highly-nonlinear patterns for more complicate and practical cases. Several works are trying to develop more suitable connection manners \cite{essd,stairnet,refinedet}, or to add more operations before combination \cite{run}.

The basic motivation of this paper is to enable the networks learn information of interest for each pyramid level in a more flexible way, given a ConvNet's feature hierarchy.
To achieve this goal, we explicitly reformulate the feature pyramid construction process as feature reconfiguration functions in a highly-nonlinear yet efficient way. To be specific, our pyramid construction employs a global attention to emphasize global information of the full image followed by a local reconfiguration to model local patch within the receptive field.
The resulting pyramid representation is capable of spreading strong semantics to all scales. Compared to previous studies including SSD and FPN-like models, our pyramid construction is more advantageous in two aspects: 1) the global-local reconfigurations are non-linear transformations, thus depicting more expressive power; 2) the pyramidal precessing for all scales are performed simultaneously and are hence more efficient than the layer-by-layer transformation (e.g. in lateral connections).

In our experiments, we compare different feature pyramid strategies within SSD architecture, and demonstrate the proposed method works more competitive in terms of accuracy and efficiency.
The main contributions of this paper are summarized as follows:
\begin{itemize}
  \item We propose the global attention and local reconfiguration for building feature pyramids to enhance multi-scale representations with semantically strong information;
  \item We compare and analysis popular feature pyramid methodologies within the standard SSD framework, and demonstrate that the proposed reconfiguration works more effective;
  \item The proposed method achieves the state-of-the-art results on standard object detection benchmarks (i.e., PASCAL VOC 2007, PASCAL VOC 2012 and MS COCO)  without losing real-time processing speed.
\end{itemize}

\section{Related work}

\textbf{Hand-engineered feature pyramids}: Prior to the widely development of deep convolutional networks, hand-craft features such as HOG \cite{hog} and SIFT \cite{sift} are popular for feature extraction. To make them scale-invariant, these features are computed over image pyramids \cite{icf,msdpm}. Several attempts have been performed on image pyramids for the sake of efficient computation \cite{fast_pyramid1,fast-fp,fast_pyramid2}. The sliding window methods over multi-scale feature pyramids are usually applied in object detection \cite{sliding1,sliding2}.

\textbf{Deep object detectors}: Benefited by the success of deep ConvNets, modern object detectors like R-CNN \cite{rcnn} and Overfeat \cite{overfeat} lead dramatic improvement for object detection. Particularly, OverFeat adopts a similar strategy to early face detectors by applying a ConvNet as the sliding window detector on image pyramids; R-CNN employs a region proposal-based strategy and classifies each scale-normalized proposal with a ConvNet.
The SPP-Net \cite{sppnet} and Fast R-CNN \cite{fastrcnn} speed up the R-CNN approach with RoI-Pooling that allows the classification layers to reuse the CNN feature maps. Since then, Faster R-CNN \cite{fasterrcnn} and R-FCN \cite{rfcn} replace the region proposal step with lightweight networks to deliver a complete end-to-end system. More recently, Redmon et al. \cite{yolo,yolo9000} propose a method named YOLO to predict bounding boxes and associate class probabilities in a single step.

\textbf{Deep feature pyramids}: To make the detection more reliable, researchers usually adopt multi-scale representations by inputting images with multiple resolutions during training and testing \cite{sppnet,resnet,ion}. Clearly, the image pyramid methods are very time-consuming as them require to compute the features on each of image scale independently and thus the ConvNet features can not be reused.
Recently, a number of approaches improve the detection performance by combining predictions from different layers in a single ConvNet.
For instance, the HyperNet \cite{hypernet} and ION \cite{ion} combine features from multiple layers before making detection. To detect objects of various sizes, the SSD \cite{ssd} spreads out default boxes of different scales to multiple layers of different resolutions within a single ConvNets. So far, the SSD is a desired choice for object detection satisfying the speed-vs-accuracy trade-off \cite{speed}.
More recently, the lateral connection (or reverse connection) is becoming popular and used in object detection \cite{dssd,fpn,ron}. The main purpose of lateral connection is to enrich the semantic information of shallow layers via the top-down pathway. In contrast to such layer-by-layer connection, this paper develops a flexible framework to integrate the semantic knowledge of multiple layers in a global-local scheme.

\section{Method}
\label{Sec:GLR}
In this section, we firstly revisit the SSD detector, then consider the recent improvements of lateral connection. Finally, we present our feature pyramid reconfiguration methodology.

\begin{figure}
\centering
\includegraphics[width=0.95\linewidth]{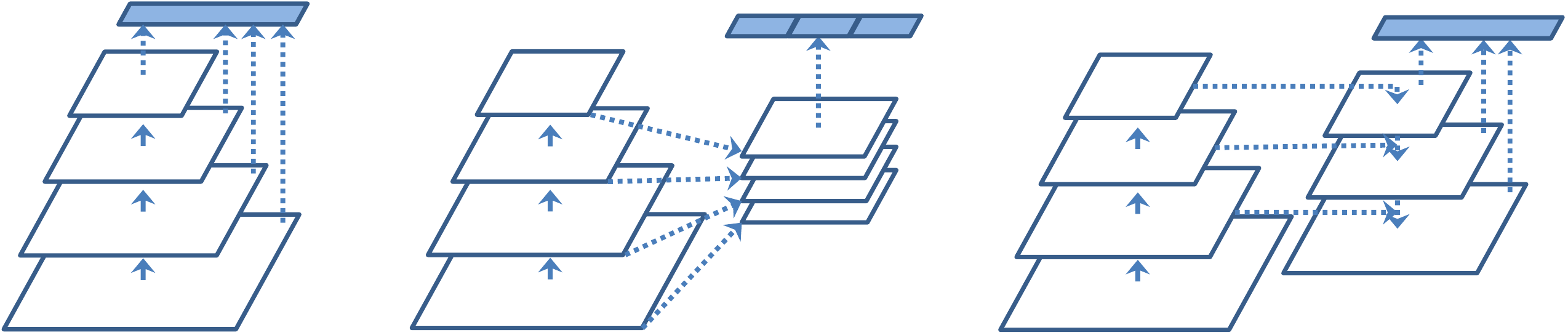}
\caption{Different feature pyramid construction frameworks. left: SSD uses pyramidal feature hierarchy computed by a ConvNet as if it is a featurized image pyramid; middle: Some object segmentation works produce final detection feature maps by directly combining features from multiple layers; right: FPN-like frameworks enforce shallow layers by top-down pathway and lateral connections.}
\label{fig:pyramid_prior}
\end{figure}

\textbf{ConvNet Feature Hierarchy}: The object detection models based on ConvNets usually adopt a backbone network (such as VGG-16, ResNets). Consider a single image $x_0$ that is passed through a convolutional network. The network comprises $L$ layers, each of which is implemented by a non-linear transformation $\mathcal{F}_l(\cdot)$, where $l$ indexes the layer. $\mathcal{F}_l(\cdot)$ is a combination transforms such as convolution, pooling, ReLU, etc. We denote the output of the $l^{th}$ layer as $x_l$. The total backbone network outputs are expressed as $X_{net} = \{x_1, x_2, ... , x_L\}$.

Without feature hierarchy, object detectors such as Faster R-CNN \cite{fasterrcnn} use one deep and semantic layer such as $x_L$ to perform object detection. In SSD \cite{ssd}, the prediction feature map sets can be expressed as
\begin{equation}
X_{pred} = \{x_{P}, x_{P+1}, \ldots, x_L\},
\end{equation}
where $P \gg 1$\footnote{For VGG-16 based model, $P = 23$ since we begin to predict from \emph{conv4\_3} layer.}. Here, the deep feature maps $x_L$ learn high-semantic abstraction. When $P < l < L$, $x_{l}$ becomes shallower thus has more low-level features. SSD uses deeper layers to detect large instances, while uses the shallow and high-resolution layers to detect small ones\footnote{Here the `small' means that the proportion of objects in the image is small, not the actual instance size. }. The high-resolution maps with limited-semantic information harm their representational capacity for object recognition. It misses the opportunity to reuse deeper and semantic information when detecting small instances, which we show is the key bottleneck to boost the performance.

\textbf{Lateral Connection}: To enrich the semantic information of shallow layers, one way is to add features from the deeper layers\footnote{When the resolutions of the two layers are not the same, usually upsample and linear projection are carried out before combination.}. Taking the FPN manner \cite{fpn} as an example, we get

\begin{gather}
\nonumber x_{L}^{'}=x_{L},~~~~~~~~~~~~~~~~~~~~~~~~~~~~~\\
\nonumber x_{L-1}^{'}=\alpha_{L-1}\cdot x_{L-1}+\beta_{L-1}\cdot x_{L},\\
 x_{L-2}^{'}=\alpha_{L-2}\cdot x_{L-2}+\beta_{L-2}\cdot x_{L-1}^{'},\\
\nonumber ~~~~~~~~~~~~~~~~~~~~~~~~~~~~~~~~~~~=\alpha_{L-2}\cdot x_{L-2}+\beta_{L-2}\alpha_{L-1}\cdot
x_{L-1}+\beta_{L-2}\beta_{L-1}\cdot x_{L},
\end{gather}
where $\alpha$, $\beta$ are weights. Without loss of generality,
\begin{equation}
x_{l}^{'}= \sum_{l=P}^{L} w_l\cdot x_{l},
\label{expend}
\end{equation}
where $w_l$ is the generated final weights for $l^{th}$ layer output after similar polynomial expansions.
Finally, the features used for detection are expressed as:
\begin{equation}
X_{pred}^{'} = \{x_{P}^{'}, x_{P+1}^{'}, \ldots, x_L^{'}\}.
\label{fpn_eq}
\end{equation}

From Eq.\ref{expend} we see that the final features $x_l^{'}$ is equivalent to the linear combination of $x_l, x_{l+1}, \ldots, x_L$. The linear combination with deeper feature hierarchy is one way to improve information of a specific shallow layer. And the linear model can achieve a good extent of abstraction when the samples of the latent concepts are linearly separable. However, the feature hierarchy for detection often lives on a non-linear manifold, therefore the representations that capture these concepts are generally highly non-linear function of the input \cite{nin,panet,senet}. It's representation power, as we show next, is not enough for the complex task of object detection.

\subsection{Deep Feature Reconfiguration}

\begin{figure}[t]
\centering
\includegraphics[width=0.6\linewidth]{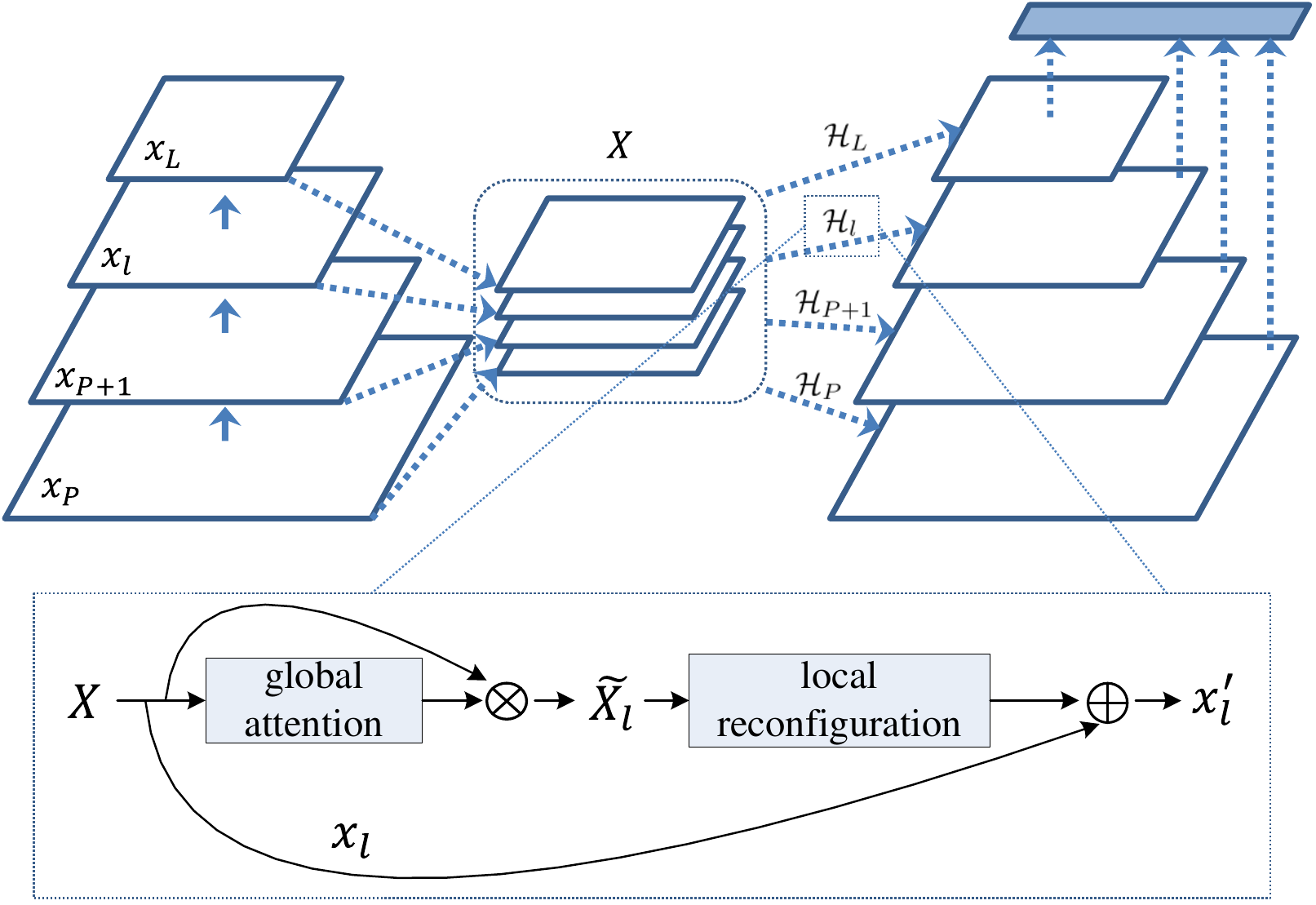}
\caption{Top: Overview of the proposed feature pyramid building networks. We firstly combine multiple feature maps, then generate features at a specific level, finally detect objects at multiple scales. Down: A building block illustrating the global attention and local reconfiguration.}
\label{fig:auto}
\end{figure}

Given the deep feature hierarchy $X = [x_{P}, x_{P+1}, \ldots, x_L]$ of a ConvNet, the key problem of object detection framework is to generate suitable features for each level of detector. In this paper, the feature generating process at $l^{th}$ level is viewed as a non-linear transformation of the given feature hierarchy (Fig. \ref{fig:auto}):
\begin{equation}
\label{trans}
x_{l}^{'} = \mathcal{H}_{l}(X)
\end{equation}
where $X$ is the feature hierarchy considered for multi-scale detection. For ease of implementation, we concatenate the multiple inputs of $\mathcal{H}_{l}(\cdot)$ in Eq.\ref{trans} into a single tensor before following transformations\footnote{For a target scale which has $W \times H$ spatial resolution, adaptive sampling is carried out before concatenation.}.

Given no priors about the distributions of the latent concepts of the feature hierarchy, it is desirable to use a universal function approximator for feature extraction of each scale. The function should also keep the spatial consistency, since the detector will activate at the corresponding locations. \emph{The final features for each level are non-linear transformations for the feature hierarchy, in which learnable parameters are shared between different spatial locations}.

In this paper, we formulate the feature transformation process $\mathcal{H}_{l}(\cdot)$ as global attention and local reconfiguration problems. Both global attention and local reconfiguration are implemented by a light-weight network so they could be embedded into the ConvNets and learned end-to-end. The global and local operations are also complementary to each other, since they deal with the feature hierarchy from different scales.


\subsubsection{Global Attention for Feature Hierarchy}
Given the feature hierarchy, the aim of the global part is to emphasise informative features and suppress less useful ones globally for a specific scale. In this paper, we apply the Squeeze-and-Excitation block \cite{senet} as the basic module.
One Squeeze-and-Excitation block consists of two steps, \emph{squeeze} and \emph{excitation}. For the $l^{th}$ level layer, the \emph{squeeze} stage is formulated as a global pooling
operation on each channel of $X$ which has $W\times H\times C$ dimensions:
\begin{equation}\label{squeese}
  z^{c}_l = \frac{1}{W \times H} \sum_{i=1}^{W}\sum_{j=1}^{H} x^c_{l}(i, j)
\end{equation}
where $x^c_{l}(i, j)$ specifies one element at $c^{th}$ channel, $i^{th}$ column and $j^{th}$ row. If there are $C$ channels in feature $X$, Eq.\ref{squeese} will generate $C$ output elements, denoted as $\textbf{z}_l$.

The excitation stage is two fully-connected layers followed by sigmoid activation with input $\textbf{z}_l$:
\begin{equation}\label{squeese}
  \textbf{s}_l = \sigma(W_l^{1}\delta(W_2^{l}\textbf{z}_l))
\end{equation}
where $\delta$ refers to the ReLU function, $\sigma$ is the sigmoid activation, $W_l^{1}\in R^{\frac{c}{r}}$ and $W_2^{2}\in R^{c}$. $r$ is set to 16 to make dimensionality-reduction. The final output of the block is obtained by rescaling the input $X$ with the activations:
\begin{equation}\label{squeese}
\tilde{\textbf{x}}^c_l = s^c_l \otimes \textbf{x}^c
\end{equation}
 then $\tilde{X}_l = [\tilde{x}^P_l, \tilde{x}^{P+1}_l, \ldots, \tilde{x}^L_l]$, $\otimes$ denotes channel-wise multiplication. More details can be referred to the SENets \cite{senet} paper.

The original SE block is developed for explicitly modelling interdependencies between channels, and shows great success in object recognition \cite{segnet}. In contrast, we apply it to emphasise channel-level hierarchy features and suppress less useful ones. By dynamically adopting conditions on the input hierarchy, SE Block helps to boost feature discriminability and select more useful information globally.

\subsubsection{Local Reconfiguration}
\begin{figure}
\centering
\includegraphics[width=0.3\linewidth]{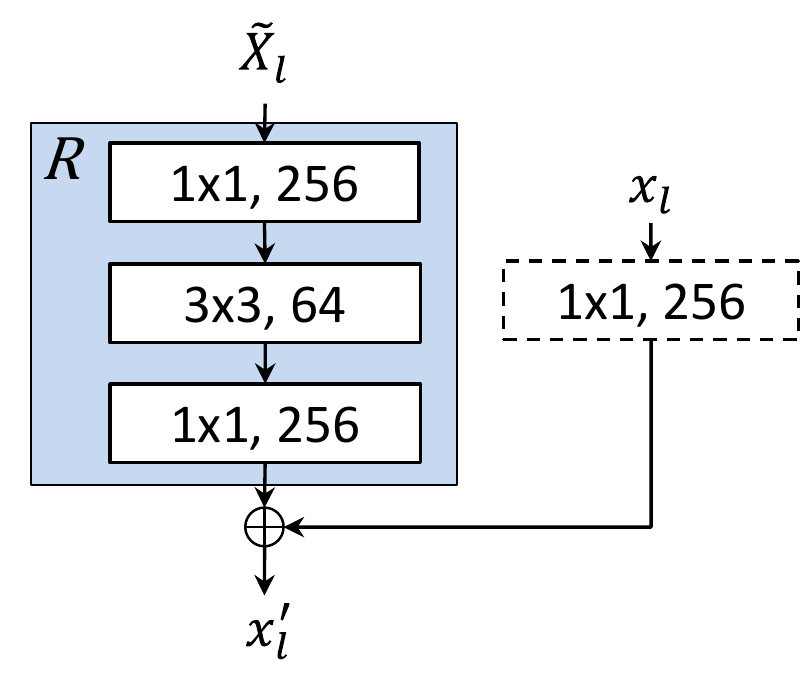}
\caption{A building block illustrating the local reconfiguration for level $l$. }
\label{fig:res}
\end{figure}

The local reconfiguration network maps the feature hierarchy patch to an output feature patch, and is shared among all local receptive fields. The output feature maps are obtained by sliding the operation over the input. In this work, we design a residual learn block as the instantiation of the micro network, which is a universal function approximator and trainable by back-propagation (Fig.\ref{fig:res}).

Formally, one local reconfiguration is defined as:

\begin{equation}\label{sss}
x_{l}^{'} = R(\tilde{X}_l)+W_l x_l
\end{equation}
where $W_l$ is a linear projection to match the dimensions\footnote{When dimensions are the same, there is no need to use it, denoted as dotted line in Fig.\ref{fig:res}}. $R(\cdot)$ represents the residual mapping that improves the semantics to be learned.

\textbf{Discussion}
A direct way to generate feature pyramids is just use the term $R(\cdot)$ in Eq.\ref{sss}. However, as demonstrated in \cite{resnet}, it is easier to optimize the residual mapping than to optimize the desired underlying mapping. Our experiments in Section \ref{discuss_res} also prove this hypothesize.

We note there are some differences between our residual learn module and that proposed in ResNets \cite{resnet}. Our hypothesize is that the semantic information is distributed among feature hierarchy and the residual learn block could select additional information by optimization. While the purpose of the residual learn in \cite{resnet} is to gain accuracy by increasing network depth. Another difference is that the input of the residual learning is the feature hierarchy, while in \cite{resnet}, the input is one level of convolutional output.


The form of the residual function $R(\cdot)$ is also flexible. In this paper, we involve a function that has three layers (Fig.\ref{fig:res}), while more layers are possible. The element-wise addition is performed on two feature maps, channel by channel. Because all levels of the pyramid use shared operations for detection,
we fix the feature dimension (numbers of channels, denoted as $d$) in all the feature maps. We set $d = 256$ in this paper and thus all layers used for prediction have 256-channel outputs.

\section{Experiments}

We conduct experiments on three widely used benchmarks: PASCAL VOC 2007, PASCAL VOC 2012 \cite{voc} and MS COCO datasets \cite{coco}. All network backbones are pretrained on the ImageNet1k classification set \cite{imagenet} and fine-tuned on the detection dataset. We use the pre-trained VGG-16 and ResNets models that are publicly available\footnote{\url{https://github.com/pytorch/vision}}. Our experiments are based on re-implementation of SSD \cite{ssd}, Faster R-CNN \cite{fasterrcnn} and Feature Pyramid Networks \cite{fpn} using PyTorch \cite{pytorch}. For the SSD framework, all layers in $\Mat{X}$ are resized to the spatial size of layer conv8\_2 in VGG and conv6\_x in ResNet-101 to keep consistency with DSSD. For the Faster R-CNN pipeline, the resized spatial size is as same as the conv4\_3 layer in both VGG and ResNet-101 backbones.

\subsection{PASCAL VOC 2007}

\textbf{Implementation details}. All models are trained on the VOC 2007 and VOC 2012 trainval sets, and tested on the VOC 2007 test set. For \emph{one-stage} SSD, we set the learn rate to $10^{-3}$ for the first 160 epochs, and decay it to $10^{-4}$ and $10^{-5}$ for another 40 and 40 epochs.
We use the default batch size 32 in training, and use VGG-16 as the backbone networks for all the ablation study experiments on the PASCAL VOC dataset.
For \emph{two-stage} Faster R-CNN experiments, we follow the training strategies introduced in \cite{fasterrcnn}. We also report the results of ResNets used in these models.

\subsubsection{Baselines} For fair comparisons with original SSD and its its feature pyramid variations, we conduct two baselines: Original SSD and SSD with feature lateral connections.
In Table \ref{table:voc_baseline}, the original SSD scores 77.5\%, which is the same as that reported in \cite{ssd}. Adding lateral connections in SSD improves results to 78.5\% (SSD+lateral). When using the global and local reconfiguration strategy proposed above, the result is improved to 79.6$\%$, which is 1.6\% better than SSD with lateral connection. In the next, we discuss the ablation study in more details.
\begin{table}\centering
\begin{center}
\caption{Effectiveness of various designs with SSD300. }
\label{table:voc_baseline}
\begin{tabular}{p{3.5cm}<{\centering}|p{2cm}<{\centering}|p{2cm}<{\centering}|p{1.5cm}<{\centering}}
method & backbone&FPS&mAP($\%$)\\
\hline
 SSD (Caffe) \cite{ssd}      &VGG-16        &46      & 77.5 \\
 SSD (ours-re)           &VGG-16    &44     & 77.5 \\
 SSD+lateral              &VGG-16   &37    & 78.5 \\
\hline
 SSD+Local only      &VGG-16        &40   &79.0\\
 SSD+Local only(no res)&VGG-16      &40       &78.6\\
 SSD+Global-Local     &VGG-16       &39.5 & \textbf{79.6} \\
\end{tabular}
\end{center}
\end{table}

\subsubsection{How important is global attention?}

In Table \ref{table:voc_baseline}, the fourth row shows the results of our model without the global attention. With this modification, we remove the global attention part and directly add local transformation into the feature hierarchy. Without global attention, the result drops to 79.0\% mAP (-0.6\%). The global attention makes the network to focus more on features with suitable semantics and helps detecting instance with variation.


\subsubsection{Comparison with the lateral connections}

Adding global and local reconfiguration to SSD improves the result to 79.6$\%$, which is 2.1\% better than SSD and 1.1\% better than SSD with lateral connection. This is because there are large semantic gaps between different levels on the bottom-up pyramid. And the global and local reconfigurations help the detectors to select more suitable feature maps. This issue cannot be simply remedied by just lateral connections. We note that only adding local reconfiguration, the result is better than lateral connection (+0.5\%).

\subsubsection{Only use the  term $R(\cdot)$}\label{discuss_res}

One way to generate the final feature pyramids is just use the term $R(\cdot)$. in Eq.\ref{sss}. Compared with residual learn block, the result drops 0.4\%. The residual learn block can avoid the gradients of the objective function to directly flow into the backbone network, thus gives more opportunity to better model the feature hierarchy.

\subsubsection{Use all feature hierarchy or just deeper layers?}

In Eq.\ref{expend}, the lateral connection only considers feature maps that are deeper (and same) than corresponding levels. To better compare our method with lateral connection, we conduct a experiment that only consider the deep layers too. Other settings are the same with the previous baselines. We find that just using deeper features drops accuracy by a small margin(-0.2\%). We think the difference is that when using the total feature hierarchy, the deeper layers also have more opportunities to re-organize its features, and has more potential for boosting results, similar conclusions are also drawn from the most recent work of PANet \cite{panet}.

\subsubsection{Accuracy vs. Speed}

We present the inference speed of different models in the third column of Table \ref{table:voc_baseline}. The speed is evaluated with batch size 1 on a machine with NVIDIA Titan X, CUDA 8.0 and cuDNN v5. Our model has a 2.7\% accuracy gain with 39.5 \emph{fps}. Compared with the lateral connection based SSD, our model shows higher accuracy and faster speed. In lateral connection based model, the pyramid layers are generated serially, thus last constructed layer considered for detection becomes the speed bottleneck ($x_{P}^{'}$ in Eq. \ref{fpn_eq}). In our design, all final pyramid maps are generated simultaneously, and is more efficient.

\subsubsection{Under Faster R-CNN pipeline}
To validate the generation of the proposed feature reconfiguration method, we conduct experiment under \emph{two-stage} Faster R-CNN pipeline. In Table \ref{table:voc_frcnn}, Faster R-CNN with ResNet-101 get mAP of 78.9\%. Feature Pyramid Networks with lateral connection improve the result to 79.8\% (+0.9\%). When replacing the lateral connection with global-local transformation, we get score of 80.6\% (+1.8\%). This result indicate that our global-and-local reconfiguration is also effective in \emph{two-stage} object detection frameworks and could improve its performance.

\begin{table}[t]\centering
\begin{center}
\caption{Effectiveness of various designs within Faster R-CNN.}
\label{table:voc_frcnn}
\begin{tabular}{p{3.5cm}<{\centering}|p{2cm}<{\centering}|p{1.5cm}<{\centering}}\noalign{\smallskip}
method & backbone &mAP($\%$)\\
\hline
 Faster \cite{fasterrcnn} &VGG-16                    & 73.2 \\
 Faster \cite{rfcn}       &ResNet-101              & 76.4 \\
 Faster(ours-re)          &ResNet-50               & 77.6 \\
 Faster(ours-re)          &ResNet-101              & 78.9 \\
\hline
 Faster+FPNs              &ResNet-50              & 78.8 \\
 Faster+FPNs              &ResNet-101              & 79.8 \\
 Faster+Global-Local      &ResNet-50              & 79.4 \\
 Faster+Global-Local      &ResNet-101              & \textbf{80.6} \\
\end{tabular}
\end{center}
\end{table}

\subsubsection{Comparison with other state-of-the-arts}

Table \ref{table:voc07} shows our results on VOC2007 test set based on SSD \cite{ssd}. Our model with $300\times 300$ achieves 79.6\% mAP, which is much better than baseline method SSD300 (77.5\%) and on par with SSD512. Enlarging the input image
to $512\times 512$ improves the result to 81.1\%. Notably our model is much better than other methods which try to include context information such as MRCNN [10] and ION \cite{ion}. When replace the backbone network from VGG-16 to ResNet-101, our model with $512\times512$ scores 82.4\% without bells and whistles, which is much better than the one-stage DSSD \cite{dssd} and two-stage R-FCN \cite{rfcn}.

\begin{table}[t]
\caption{PASCAL VOC 2007 test detection results. All models are trained with 07+12 (07 trainval + 12 trainval). The entries with the best APs for each object category are bold-faced.}
\label{table:voc07}
\resizebox{\textwidth}{!}{
\begin{tabular}{p{2cm}|p{1.7cm}<{\centering}|p{1.3cm}<{\centering}|p{0.75cm}<{\centering}p{0.75cm}<{\centering}p{0.75cm}<{\centering}p{0.75cm}<{\centering}p{0.75cm}<{\centering}
p{0.75cm}<{\centering}p{0.75cm}<{\centering}p{0.75cm}<{\centering}p{0.75cm}<{\centering}p{0.75cm}<{\centering}p{0.75cm}<{\centering}p{0.75cm}<{\centering}p{0.75cm}<{\centering}
p{0.75cm}<{\centering}p{1cm}<{\centering}p{0.75cm}<{\centering}p{0.75cm}<{\centering}p{0.75cm}<{\centering}p{0.75cm}<{\centering}p{0.75cm}<{\centering}}
\noalign{\smallskip}
 Method   & backbone    & mAP($\%$) & aero  &bike   &bird   &boat   &bottle &bus &car &cat &chair &cow &table &dog &horse &mbike &person &plant &sheep &sofa &train &tv\\
\hline
 Faster\cite{fasterrcnn}    & VGG-16    &73.2   &76.5   &79.0   &70.9   &65.5   &52.1   &83.1   &84.7   &86.4   &52.0   &81.9   &65.7   &84.8   &84.6   &77.5   &76.7   &38.8   &73.6  &73.9  &83.0   &72.6 \\
 ION\cite{ion}              & VGG-16    &76.5   &79.2   &79.2   &77.4   &69.8   &55.7   &85.2   &84.2   &89.8   &57.5   &78.5   &73.8   &87.8   &85.9   &81.3   &75.3   &49.7   &76.9  &74.6  &85.2   &82.1 \\
 MRCNN\cite{mrcnn}          & VGGNet    &78.2   &80.3   &84.1   &78.5   &70.8   &68.5   &88.0   &85.9   &87.8   &60.3   &85.2   &73.7   &87.2   &86.5   &85.0   &76.4   &48.5   &76.3  &75.5  &85.0   &81.0 \\
 Faster\cite{fasterrcnn}    & ResNet-101 &76.4   &79.8   &80.7   &76.2   &68.3   &55.9   &85.1   &85.3   &89.8   &56.7   &87.8   &69.4   &88.3   &88.9   &80.9   &78.4   &41.7   &78.6  &79.8  &85.3   &72.0 \\
 R-FCN\cite{rfcn}           & ResNet-101 &80.5   &79.9   &87.2   &81.5   &72.0   &\textbf{69.8}   &86.8   &88.5   &89.8   &\textbf{67.0}   &88.1   &74.5   &89.8   &\textbf{90.6}   &79.9   &81.2   &53.7   &81.8   &\textbf{81.5}  &85.9   &79.9 \\
  SSD300\cite{ssd}           & VGG-16    &77.5   &79.5   &83.9   &76.0   &69.6   &50.5   &87.0   &85.7   &88.1   &60.3   &81.5   &77.0   &86.1   &87.5   &83.9  &79.4   &52.3   &77.9   &79.5   &87.6   &76.8\\
 SSD512\cite{ssd}           & VGG-16    &79.5   &84.8   &85.1   &81.5   &73.0   &57.8   &87.8   &88.3   &87.4   &63.5   &85.4   &73.2   &86.2   &86.7   &83.9   &82.5   &55.6   &81.7   &79.0  &86.6   &80.0\\
 StairNet\cite{stairnet}    & VGG-16    &78.8   &81.3   &85.4   &77.8   &72.1   &59.2   &86.4   &86.8   &87.5   &62.7   &85.7   &76.0   &84.1   &88.4   &86.1   &78.8   &54.8   &77.4   &79.0  &88.3   &79.2\\
 RON320\cite{ron}           & VGG-16    &76.6   &79.4   &84.3   &75.5   &69.5   &56.9   &83.7   &84.0   &87.4   &57.9   &81.3   &74.1   &84.1   &85.3   &83.5   &77.8   &49.2   &76.7  &77.3  &86.7   &77.2 \\
 DSSD321\cite{dssd}         & ResNet-101 &78.6   &81.9   &84.9   &80.5   &68.4   &53.9   &85.6   &86.2   &88.9   &61.1   &83.5   &78.7   &86.7   &88.7   &86.7   &79.7   &51.7   &78.0   &80.9  &87.2  &79.4\\
 DSSD513\cite{dssd}         & ResNet-101 &81.5   &86.6   &86.2   &\textbf{82.6}   &74.9   &62.5   &\textbf{89.0}   &88.7   &88.8   &65.2   &87.0   &78.7   &88.2   &89.0   &87.5   &83.7   &51.1   &86.3  &81.6  &85.7   &\textbf{83.7}\\
 \hline
 Ours300                    & VGG-16    &79.6   &84.5   &85.5   &77.2   &72.1   &53.9   &87.6   &87.9   &89.4   &63.8   &86.1   &76.1   &87.3   &88.8   &86.7   &80.0   &54.6   &80.5   &81.2  &88.9   &80.2 \\
 Ours512                    & VGG-16    &81.1   &90.0   &87.0   &79.9   &75.1   &60.3   &88.8   &\textbf{89.6}   &89.6   &65.8   &\textbf{88.4}   &79.4   &87.5   &90.1   &85.6   &81.9   &54.8   &79.0   &80.8  &87.2   &79.9 \\
 Ours300                    & ResNet-101 &80.2	&89.3	&84.9	&79.9	&\textbf{75.6}	&55.4	&88.2	&88.6	&88.6	&63.3	&87.9	&78.8	&87.3	&87.7	&85.5	&80.5	&55.4	&81.1	&79.6	&87.8	&78.5 \\
 Ours512                    & ResNet-101 &\textbf{82.4}	&\textbf{92.0}	&\textbf{88.2}	&81.1	&71.2	&65.7	&88.2	&87.9	&\textbf{92.2}	&65.8	&86.5	&\textbf{79.4}	&\textbf{90.3}	&90.4	&89.3	&\textbf{88.6}	&\textbf{59.4}	&\textbf{88.4}	&75.3	&\textbf{89.2}	&78.5 \\
\end{tabular}
}
\end{table}

\begin{figure}[t]
\centering
\begin{minipage}[t]{0.3\linewidth}
\centering
\includegraphics[width=0.95\linewidth]{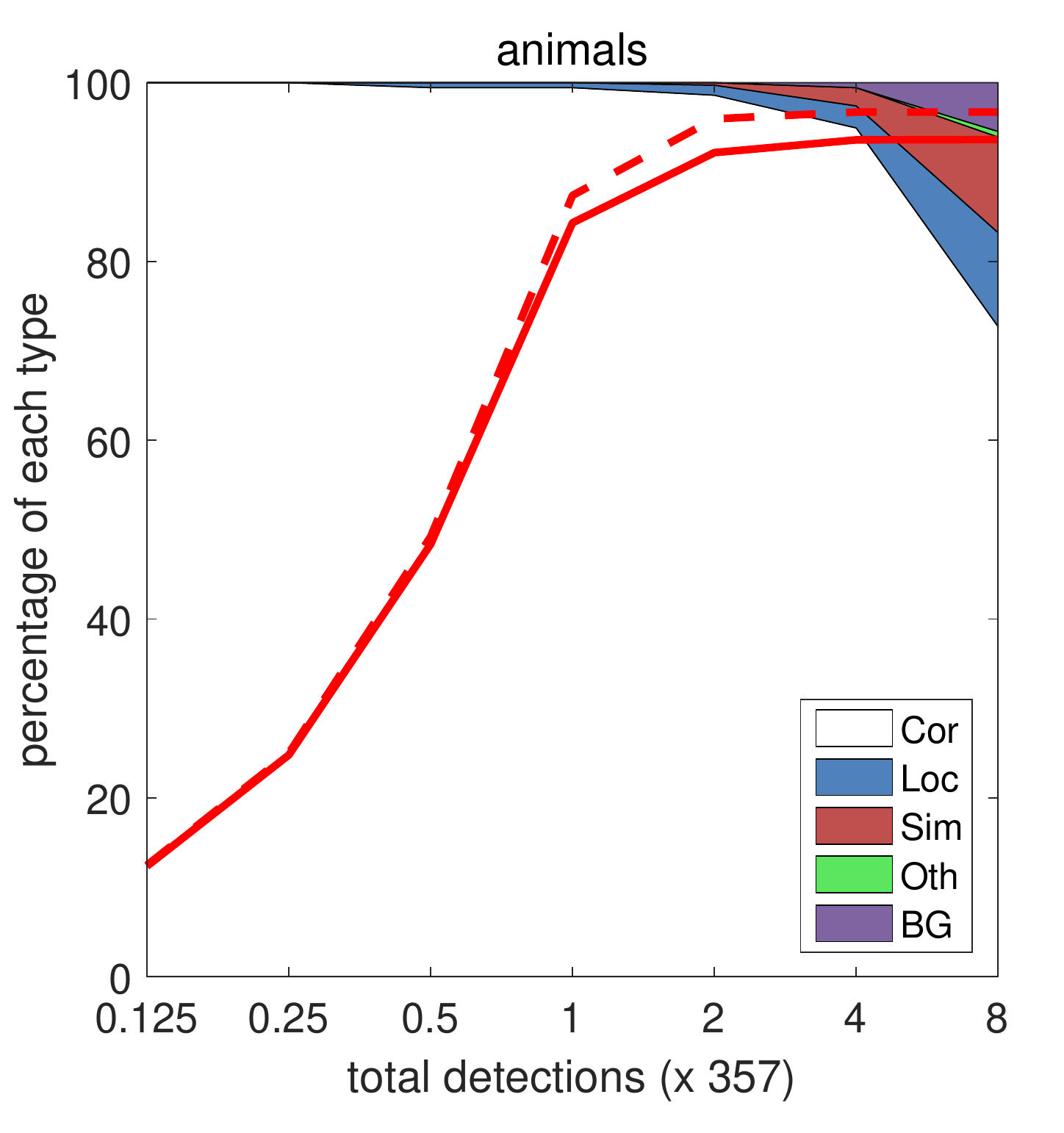}
\end{minipage}
\begin{minipage}[t]{0.3\linewidth}
\centering
\includegraphics[width=0.95\linewidth]{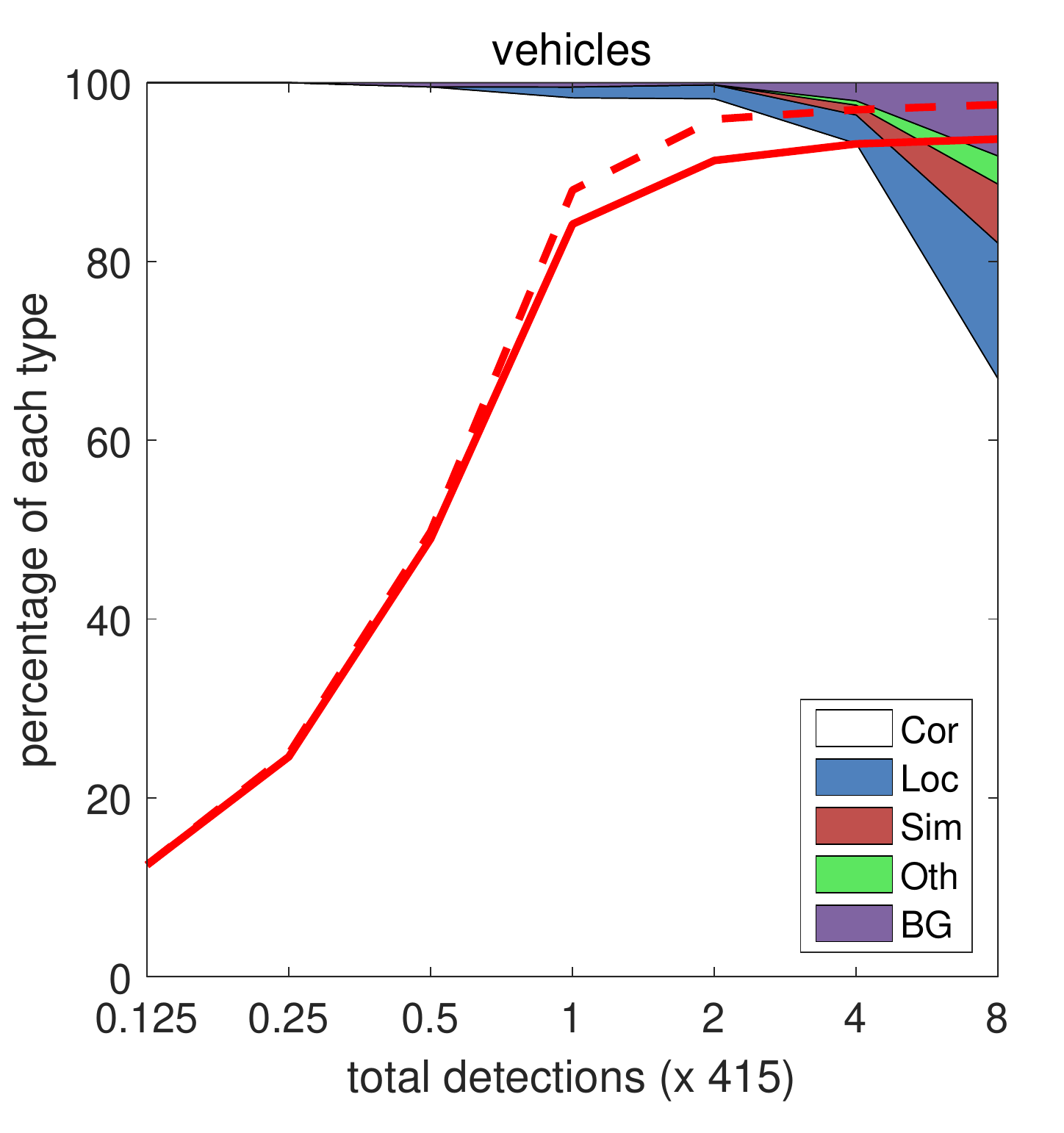}
\end{minipage}
\begin{minipage}[t]{0.3\linewidth}
\centering
\includegraphics[width=0.95\linewidth]{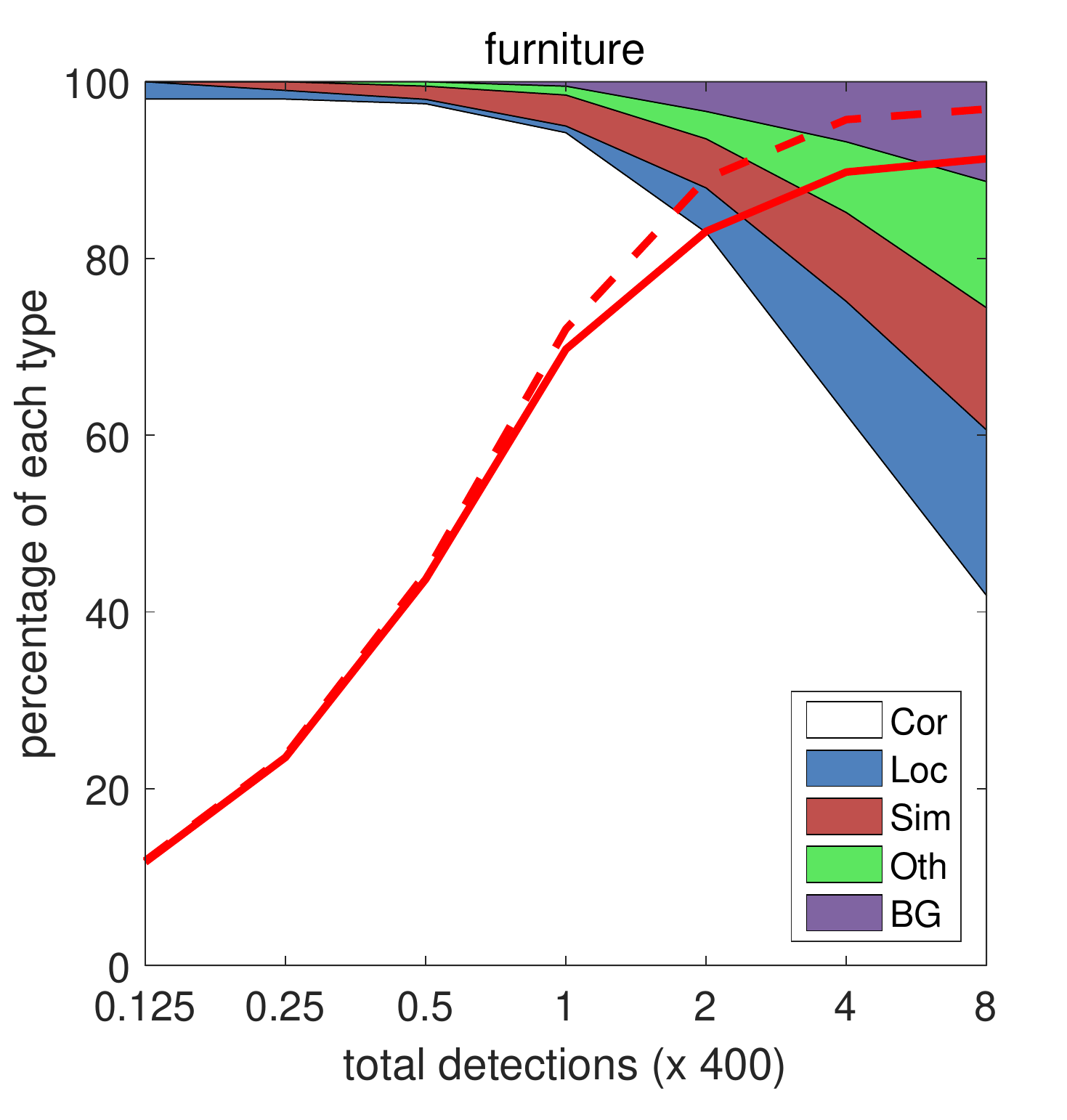}
\end{minipage}
\caption{Visualization of performance for our model with VGG-16 and $300\times300$ input resolution on animals, vehicles, and furniture from VOC2007 test. The Figures show the cumulative fraction of detections that are correct (Cor) or false positive due to poor localization (Loc), confusion with similar categories (Sim), with others (Oth), or with background (BG). The solid red line reflects the change of recall with the `strong' criteria (0.5 jaccard overlap) as the number of detections increases. The dashed red line uses the `weak' criteria (0.1 jaccard overlap).}
\label{det_error}
\end{figure}

To understand the performance of our method in more detail, we use the detection analysis tool from \cite{error}. Figure \ref{det_error} shows that our model can detect various object categories with high quality. The recall is higher than 90\%, and is much higher with the `weak' (0.1 jaccard overlap) criteria.

\subsection{PASCAL VOC 2012}

For VOC2012 task, we follow the setting of VOC2007 and with a few differences described here. We use 07++12 consisting of VOC2007 trainval, VOC2007 test, and
VOC2012 trainval for training and VOC2012 test for testing. We see the same performance trend as we observed on VOC 2007 test. The results, as shown in Table \ref{table:voc12}, demonstrate the effectiveness of our models. Compared with SSD \cite{ssd} and other variants, the proposed network is significantly better(+2.7\% with $300\times 300$).

\begin{table}
\caption{PASCAL VOC 2012 test detection results. All models are trained with 07++12 (07 trainval+test + 12 trainval). The entries with the best APs for each object category are bold-faced}
\label{table:voc12}
\resizebox{\textwidth}{!}{
\begin{tabular}{p{2cm}|p{1.7cm}<{\centering}|p{1.3cm}<{\centering}|p{0.75cm}<{\centering}p{0.75cm}<{\centering}p{0.75cm}<{\centering}p{0.75cm}<{\centering}p{0.75cm}<{\centering}
p{0.75cm}<{\centering}p{0.75cm}<{\centering}p{0.75cm}<{\centering}p{0.75cm}<{\centering}p{0.75cm}<{\centering}p{0.75cm}<{\centering}p{0.75cm}<{\centering}p{0.75cm}<{\centering}
p{0.75cm}<{\centering}p{1cm}<{\centering}p{0.75cm}<{\centering}p{0.75cm}<{\centering}p{0.75cm}<{\centering}p{0.75cm}<{\centering}p{0.75cm}<{\centering}}
\noalign{\smallskip}
 Method   & network    & mAP($\%$) & aero  &bike   &bird   &boat   &bottle &bus &car &cat &chair &cow &table &dog &horse &mbike &person &plant &sheep &sofa &train &tv\\
\hline
 Faster\cite{fasterrcnn}    & ResNet-101 &73.8   &86.5   &81.6   &77.2   &58.0   &51.0   &78.6   &76.6   &93.2   &48.6   &80.4   &59.0   &92.1   &85.3   &84.8   &80.7   &48.1   &77.3  &66.5  &84.7   &65.6 \\
 R-FCN\cite{rfcn}           & ResNet-101 &77.6   &86.9   &83.4   &\textbf{81.5}   &63.8   &62.4   &81.6   &81.1   &93.1   &58.0   &83.8   &60.8   &92.7   &86.0   &84.6   &84.4   &59.0   &80.8  &68.6  &86.1   &72.9 \\
 ION\cite{ion}              & VGG-16    &76.4   &87.5   &84.7   &76.8   &63.8   &58.3   &82.6   &79.0   &90.9   &57.8   &82.0   &64.7   &88.9   &86.5   &84.7   &82.3   &51.4   &78.2  &69.2  &85.2   &73.5 \\
 SSD300\cite{ssd}           & VGG-16    &75.8   &88.1   &82.9   &74.4   &61.9   &47.6   &82.7   &78.8   &91.5   &58.1   &80.0   &64.1   &89.4   &85.7   &85.5   &82.6   &50.2   &79.8  &73.6  &86.6   &72.1\\
 SSD512\cite{ssd}           & VGG-16    &78.5   &90.0   &85.3   &77.7   &64.3   &58.5   &85.1   &84.3   &92.6   &61.3   &83.4   &65.1   &89.9   &88.5   &88.2   &85.5   &54.4   &82.4  &70.7  &87.1   &75.6\\
 DSSD321\cite{dssd}         & ResNet-101 &76.3   &87.3   &83.3   &75.4   &64.6   &46.8   &82.7   &76.5   &92.9   &59.5   &78.3   &64.3   &91.5   &86.6   &86.6   &82.1   &53.3   &79.6  &75.7  &85.2   &73.9\\
 DSSD513\cite{dssd}         & ResNet-101 &80.0   &92.1   &86.6   &80.3   &68.7   &58.2   &84.3   &85.0   &\textbf{94.6}   &63.3   &85.9   &65.6   &\textbf{93.0}   &88.5   &87.8   &86.4   &57.4   &85.2  &73.4  &87.8   &76.8\\
 YOLOv2\cite{yolo9000}      &Darknet-19 &75.4   &86.6	&85.0	&76.8	&61.1	&55.5	&81.2	&78.2	&91.8	&56.8	&79.6	&61.7	&89.7	&86.0	&85.0	&84.2	&51.2	&79.4	&62.9	&84.9	&71.0\\
 DSOD\cite{dsod}            &DenseNet   &76.3   &89.4	&85.3	&72.9	&62.7	&49.5	&83.6	&80.6	&92.1	&60.8	&77.9	&65.6	&88.9	&85.5	&86.8	&84.6	&51.1	&77.7	&72.3	&86.0	&72.2\\
 RUN300\cite{run}           &VGG-16     &77.1	&88.2	&84.4	&76.2	&63.8	&53.1	&82.9	&79.5	&90.9	&60.7	&82.5	&64.1	&89.6	&86.5	&86.6	&83.3	&51.5	&83.0	&74.0	&87.6	&74.4\\
 RUN512\cite{run}           &VGG-16     &79.8	&\textbf{90.0}	&87.3	&80.2	&67.4	&62.4	&84.9	&\textbf{85.6}	&92.9	&61.8	&84.9	&66.2	&90.9	&89.1	&88.0	&86.5	&55.4	&\textbf{85.0}	&72.6	&87.7	&76.8\\
 StairNet\cite{stairnet}    &VGG-16     &76.4   &87.7   &83.1   &74.6   &64.2   &51.3   &83.6   &78.0   &92.0   &58.9   &81.8   &\textbf{66.2}   &89.6   &86.0   &84.9   &82.6   &50.9   &80.5 &71.8  &86.2   &73.5\\
 \hline
 Ours300                    & VGG-16    &77.5   &89.5   &85.0   &77.7   &64.3   &54.6   &81.6   &80.0   &91.6   &60.0   &82.5   &64.7   &89.9   &85.4   &86.1   &84.1   &53.2   &81.0   &74.2  &87.9   &75.9 \\
 Ours512                    & VGG-16    &80.0   &89.6   &\textbf{87.4 }  &80.9   &68.3   &61.0   &83.5   &83.9   &92.4   &63.8   &85.9   &63.9   &89.9   &89.2   &88.9   &86.2   &56.3   &84.4   &75.5  &89.7   &78.5 \\
 Ours300                    & ResNet-101 &78.7   &89.4   &85.7   &80.2   &65.1   &58.6   &84.3   &81.8   &91.9   &63.6   &84.2   &65.6   &89.6   &85.9   &86.0   &85.0   &54.4   &81.9   &\textbf{75.9}  &87.8   &77.5 \\
 Ours512                    & ResNet-101 &\textbf{81.1}   &87.4   &85.7   &81.4   &\textbf{71.1}   &\textbf{64.3}   &\textbf{85.1}   &84.8   &92.2   &\textbf{66.3}   &\textbf{87.6}   &66.1   &90.3   &\textbf{90.1}   &\textbf{89.6}   &\textbf{87.2}   &\textbf{60.0}   &84.4   &75.7  &\textbf{89.7}   &\textbf{80.1} \\
\end{tabular}
}
\end{table}

Compared with DSSD with ResNet-101 backbone, our model gets similar results with VGG-16 backbone. The most recently proposed RUN \cite{run} improves the results of SSD with skip-connection and unified prediction. The method add several residual blocks to improve the non-linear ability before prediction. Compared with RUN, our model is more direct and with better detection performance. Our final result using ResNet-101 scores 81.1\%, which is much better than the state-of-the-art methods.

\subsection{MS COCO}

To further validate the proposed framework on a larger and more challenging dataset, we conduct experiments on MS COCO \cite{coco} and report results from test-dev evaluation server. The evaluation metric of MS COCO dataset is different from PASCAL VOC. The average mAP over different IoU thresholds, from 0.5 to 0.95 (written as 0.5:0.95) is the overall performance of methods. We use the 80k training images and 40k validation images \cite{coco} to train our model, and validate the performance on the test-dev dataset which contains 20k images. For ResNet-101 based models, we set batch-size as 32 and 20 for $320\times320$ and $512\times512$ model separately, due to the memory issue.

\begin{table}
\begin{center}
\begin{tabular}{p{2.1cm}|p{2cm}<{\centering}|p{2.2cm}<{\centering}|p{2cm}<{\centering}|p{0.8cm}<{\centering}p{0.8cm}<{\centering}c}
\multirow{2}{*}{method}&\multirow{2}{*}{train Data}&\multirow{2}{*}{input size}&\multirow{2}{*}{network}&\multicolumn{3}{c}{Average Precision} \\
 & & & &0.5 & 0.75 &0.5:0.95 \\
\hline
\emph{two-stage} &&&\\
OHEM++\cite{ohem}               &trainval       &$\sim1000\times600$       &VGG-16    &45.9& 26.1 & 25.5\\
Faster\cite{fasterrcnn}   &trainval       &$\sim1000\times600$       &VGG-16    &42.7& - & 21.9\\
R-FCN\cite{rfcn}                &trainval       &$\sim1000\times600$       &ResNet-101  &51.9 & - &29.9 \\
CoupleNet\cite{couplenet}        &trainval35k    &$\sim1000\times600$    &ResNet-101 &\textbf{54.8} &37.2 &34.4\\
\emph{one-stage} &&&\\
SSD300\cite{ssd}                &trainval35k    &$300\times300 $   &VGG-16    &43.1& 25.8 & 25.1\\
SSD512\cite{ssd}                &trainval35k    &$512\times512$    &VGG-16    &48.5& 30.3 & 28.8\\
SSD513\cite{dssd}               &trainval35k    &$513\times513$    &ResNet-101    &50.4& 33.1 & 31.2\\
DSSD321\cite{dssd}              &trainval35k    &$321\times321$    &ResNet-101 &46.1 &29.2 &28.0 \\
DSSD513\cite{dssd}              &trainval35k    &$513\times513$    &ResNet-101 &53.3 &35.2 &33.2\\
RON320\cite{ron}                &trainval       &$320\times320 $   &VGG-16    &47.5& 25.9 & 26.2\\
YOLOv2\cite{yolo9000}           &trainval35k    &$544\times544 $    &DarkNet-19 &44.0 &19.2 &21.6 \\
RetinaNet\cite{focal}        &trainval35k    &$500\times500 $    &ResNet-101 &53.1 &36.8 &34.4\\
\hline
Ours300                         &trainval       &$300\times300 $    &VGG-16    &48.2   &29.1   &28.4\\
Ours512                         &trainval       &$512\times512$    &VGG-16    &50.9   &32.2   &31.5\\
Ours300                         &trainval       &$300\times300 $    &ResNet-101    &50.5   &32.0   &31.3\\
Ours512                         &trainval       &$512\times512$    &ResNet-101    &54.3   &\textbf{37.3}   &\textbf{34.6}\\
\end{tabular}
\end{center}
\caption{MS COCO test-dev2015 detection results.}
\label{time_comp}
\end{table}

With the standard COCO evaluation metric, SSD300 scores 25.1\% AP, and our model improves it to 28.4\% AP (+3.3\%), which is also on par with DSSD with ResNet-101 backbone(28.0\%). When change the backbone to ResNet-101, our model gets 31.3\% AP, which is much better than the DSSD321(+3.3\%). The accuracy of our model can be improved to 34.6\% by using larger input size of $512\times512$, which is also better than the most recently proposed RetinaNet \cite{focal} that adds lateral connection and focal loss for better object detection.

Table~\ref{coco_scale} reports the multi-scale object detection results of our method under SSD framework using ResNet-101 backbone. It is observed that our method achieves better detection accuracies than SSD and DSSD for the objects of all scales.

\begin{table}
\begin{center}
\begin{tabular}{p{2cm}<{\centering}|p{1cm}<{\centering}p{1cm}<{\centering}p{1cm}<{\centering}p{1cm}<{\centering}}
Methods&$AP_{s}$ & $AP_{m}$ &$AP_{l}$&$AP$ \\
 \hline
 SSD513	&10.2	&34.5   &49.8  &31.2\\
 DSSD513	&13.0	&35.4   &51.1  &33.2\\
 Ours512    		&14.7		&38.1   &51.9  &34.6\\
 \end{tabular}
\end{center}
\caption{MS COCO test-dev2015 detection results on small ($AP_s$), medium ($AP_m$) and large ($AP_l$) objects.}
\label{coco_scale}
\end{table}

\section{Conclusions}

A key issue for building feature pyramid representations under a ConvNet is to reconfigure and reuse the feature hierarchy. This paper deal with this problem with global-and-local transformations. This representation allows us to explicitly model the feature reconfiguration process for the specific scales of objects. We conduct extensive experiments to compare our method to other feature pyramid variations. Our study suggests that despite the strong representations of deep ConvNet, there is still room and potential to building better pyramids to further address multiscale problems.

\textbf{Acknowledgement}
This work was jointly supported by the National Science Fundation of China(NSFC) and the German Research Foundation(DFG) joint project NSFC 61621136008/DFG TRR-169 and the National Natural Science Foundation of China(Grant No: 61327809,61210013).

 \begin{figure}
 \centering
 \includegraphics[width=0.85\linewidth,trim=0 2800 0 0, clip]{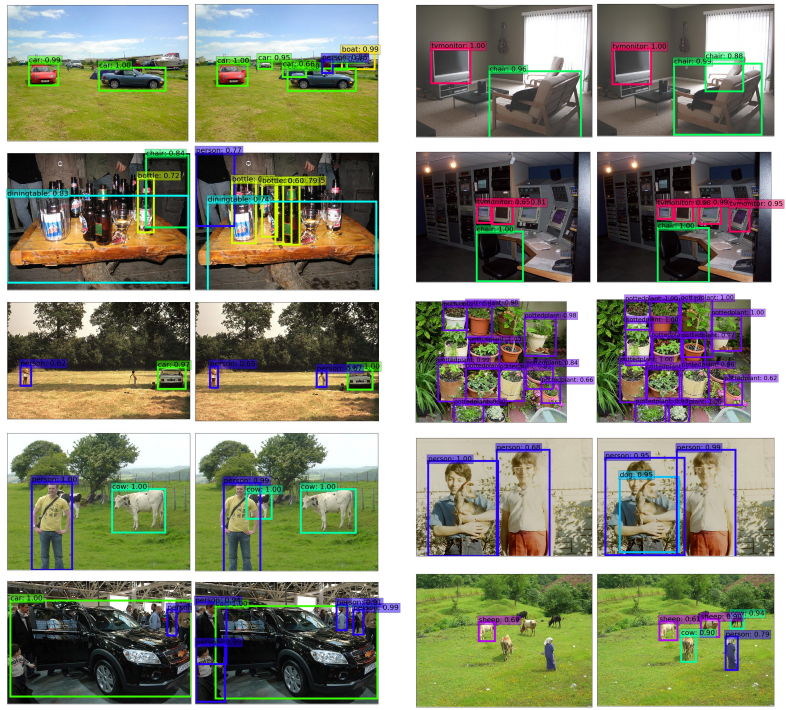}
 \caption{Qualitative detection examples on VOC 2007 test set with SSD300 (77.5\% mAP) and Ours-300 (79.6\% mAP) models. For each pair, the left is the result of
 SSD and right is the result of ours. We show detections with scores higher than 0.6. Each color corresponds to an object category in that image.}
 \label{fig:vis_voc}
 \end{figure}

%
%
%

\bibliographystyle{splncs04}
\bibliography{egbib}

\end{document}